\pdfoutput=1
\documentclass[11pt]{article}

\usepackage[preprint]{acl}

\usepackage{times}
\usepackage{latexsym}
\usepackage[T1]{fontenc}
\usepackage[utf8]{inputenc}
\usepackage{microtype}
\usepackage{inconsolata}
\usepackage{graphicx}
\usepackage{multirow}
\usepackage{colortbl}
\title{Defining Boundaries: The Impact of Domain Specification on Cross-Language and Cross-Domain Transfer in Machine Translation}

\author{Lia Shahnazaryan \and Meriem Beloucif \\
Uppsala University \\ Uppsala, Sweden \\ 
\texttt{lia.shahnazaryan.9301@student.uu.se} \\ \texttt{meriem.beloucif@lingfil.uu.se}\\}

\begin{document}
\maketitle
\begin{abstract}
Recent advancements in neural machine translation (NMT) have revolutionized the field, yet the dependency on extensive parallel corpora limits progress for low-resource languages and domains. Cross-lingual transfer learning offers a promising solution by utilizing data from high-resource languages but often struggles with in-domain NMT. This paper investigates zero-shot cross-lingual domain adaptation for NMT, focusing on the impact of domain specification and linguistic factors on transfer effectiveness. Using English as the source language and Spanish for fine-tuning, we evaluate multiple target languages, including Portuguese, Italian, French, Czech, Polish, and Greek. We demonstrate that both language-specific and domain-specific factors influence transfer effectiveness, with domain characteristics playing a crucial role in determining cross-domain transfer potential. We also explore the feasibility of zero-shot cross-lingual cross-domain transfer, providing insights into which domains are more responsive to transfer and why. Our results show the importance of well-defined domain boundaries and transparency in experimental setups for in-domain transfer learning.  
\end{abstract}
\section{Introduction}
Advancements in neural machine translation (NMT) have transformed the field, but these systems often require large parallel corpora, which are scarce for low-resource languages. Cross-lingual transfer learning has emerged as a solution, leveraging high-resource language data to improve translation quality for low-resource languages. However, a critical limitation is the pre-training on heterogeneous data, which hampers the translation of specialized texts due to a mismatch between training data and the target domain. Domain adaptation mitigates this by adjusting NMT models to specific domains, enhancing translation performance for specialized content such as legal or medical texts. Despite advancements, the intersection of cross-lingual transfer learning and domain adaptation—specifically zero-shot cross-lingual domain adaptation—remains under-explored. 

In this paper, we investigate zero-shot cross-lingual domain adaptation for NMT, integrating transfer learning across languages with domain adaptation. The objective is to fine-tune multilingual pre-trained NMT models with domain-specific data from a resource-rich language pair, capturing domain-specific knowledge and transferring it to low-resource languages within the same domain. We focus our study on the following questions:
\begin{enumerate}
    \item \textbf{Enhancement of Domain-Specific Quality:} Evaluating whether the domain-specific quality of machine translation (MT) output for one language pair can be improved by fine-tuning the model on domain-relevant data from another language pair.
    \item \textbf{Transferability of Domains:} Identifying the transferable and non-transferable domains within the scope of zero-shot cross-lingual domain adaptation for NMT.
    \item \textbf{Influence of Language-Specific vs. Domain-Specific Factors:} Analyzing the relative influence of language-specific and domain-specific factors on the effectiveness of zero-shot cross-lingual domain adaptation.
\end{enumerate}
\begin{figure}[ht]
\centering
\includegraphics[width=\columnwidth]{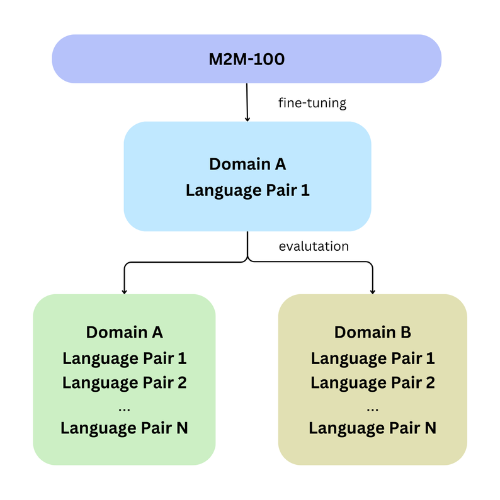}
    \caption{Illustration of the zero-shot cross-lingual domain adaptation and zero-shot cross-lingual cross-domain transfer setups.}
    \label{fig:setup}
\end{figure}
Languages explored include English as the source, Spanish for fine-tuning, and Portuguese, Italian, French, Czech, Polish, and Greek as evaluation targets, representing varying linguistic similarities. Results demonstrate that domain-specific translation quality improves through zero-shot cross-lingual domain adaptation, with specialized domains (e.g., medical, legal, IT) benefiting more than mixed domains (e.g., movie subtitles, TED talks). The study underscores the critical role of well-defining domain data to effectively transfer domain-specific knowledge across languages and domains.
\section{Background and Related Work}
\subsection{Transfer Learning Across Languages}
Building on the advancements in NMT, particularly the Transformer architecture \citep{transformer1}, transfer learning techniques have gained significant interest in recent years. The Transformer’s self-attention mechanism allows for effective modeling of long-range dependencies, leading to state-of-the-art performance in machine translation (MT) tasks. However, the success of these models is heavily reliant on the availability of large-scale parallel corpora, which are often scarce for low-resource languages. Cross-lingual transfer learning provides a promising solution to this data scarcity challenge by leveraging knowledge acquired from high-resource languages. The common approach uses multilingual pre-trained models like mT5 \citep{mt5}, mBERT \citep{mbert} and XLM-R \citep{xlmr} that are initially trained on large multilingual corpora to capture cross-lingual representations. These models can then be fine-tuned on limited parallel data for low-resource languages, transferring knowledge from the high-resource languages present in the pre-training data \citep{multi1, multi2}. Importantly, the efficiency depends on the linguistic proximity between the languages involved \citep{lang_relatedness}. Extending further, recent work explores zero-shot translation capabilities for unseen language pairs when no parallel data exists, relying solely on multilingual pre-training \citep{zero_shot_solution, zero_shot3}. Approaches include pivoting through high-resource languages \citep{zero_shot_pivot}, modifying architectures to build universal encoders that map diverse languages into shared representations \citep{zero_shot_solution2}, and auxiliary training objectives encouraging cross-lingual similarity \citep{zero_shot_solution1}. While promising, zero-shot translation remains challenging due to linguistic dissimilarities between languages and the model's ability to generalize across diverse language pairs \citep{transferability_review, transferability_problem}.
\subsection{Domain Adaptation}
A separate key challenge in NMT is domain adaptation, as general-purpose NMT systems struggle to effectively translate specialized domains like legal or medical texts due to vocabulary and stylistic mismatches from their training data \citep{domain1}. A crucial aspect is how domains are defined. The conventional view defines a domain as ``a corpus from a specific source, which may differ from other domains in terms of topic, genre, style, level of formality, etc.'' \citep{domain1}. \citet{domain2van} provide a more comprehensive view, defining a domain as a combination of provenance, topic, and genre, where provenance refers to the source of a given text, the topic pertains to the subject matter, and genre encompasses the function, register, syntax, and style of the text, as defined by \citet{genre}. Of these, topic and genre are regarded as the most critical complementary features for characterizing a domain effectively \citep{domain3danielle}. \citet{plank2barbara016non}, on the other hand, argues that topic and genre may not fully capture all domain factors, suggesting other aspects like sentence type, language, etc. Despite this, much of the research on domain adaptation in MT has mainly focused on genre as the primary domain differentiator, constructing experiments around datasets like OpenSubtitles\footnote{\url{http://www.opensubtitles.org/}} or TED \citep{datamodelcombo1, m4adapter, wrongdomaindef2},\footnote{\url{https://www.ted.com}} which offer data within the same genre while overlooking even topic specifics. However, a more comprehensive approach should account for both topic and genre, as well as other domain-specific language patterns that may impact translation quality.
\subsection{Zero-Shot Cross-Lingual Domain Adaptation}
Building on transfer learning across languages and domain adaptation, zero-shot cross-lingual domain adaptation tackles adapting multilingual NMT to specialized domains for languages with limited parallel in-domain data. The approach leverages a multilingual pre-trained NMT model fine-tuned on domain-specific data from a high-resource language pair, enabling it to capture domain knowledge. This adapted model can then translate target languages within the same domain, transferring domain knowledge in a zero-shot manner. The effectiveness of this approach depends on several influencing factors, including the linguistic proximity between the pivot and target languages, the nature and complexity of the domain itself, as well as the composition of the initial general-purpose pre-training data. Moreover, as the model transitions from general pre-training to specialized fine-tuning, there is also a risk of catastrophic forgetting where previously learned general knowledge is overwritten \citep{domain3danielle}. Approaches like embedding freezing \citep{zero_shot_domain} have proved effective in mitigating this issue. While \citet{zero_shot_domain} demonstrated the feasibility of zero-shot cross-lingual domain adaptation for the medical domain, comprehensive analysis across diverse languages and domains is still needed to understand these influencing factors and their relative impacts fully.
\section{Experimental Setup}
\label{setup}
\subsection{Domains, Datasets, and Languages}
We curate data from six different domains, encompassing both specialized areas with well-defined topics and genres, as well as more mixed domains with distinct genres but diverse topics. The three main specialized domains under focus are medical (documents related to medicinal products and their use), legal (European Union laws), and information technology (IT) (localization documents and technical user manuals). Additionally, we include two domains (movie subtitles and TED talks) that do not strictly adhere to the conventional definition of a domain, as they exhibit distinct genres but lack a specific topical focus. These two are primarily included for experimental purposes to understand the importance of domain specification in domain adaptation for NMT. The sixth domain we include is a general-purpose domain (sample of Wikipedia articles and online newspapers) to assess whether improvements in translation quality are consistent across all domains or specific to those that diverge significantly from the pre-training data (see Table \ref{tab:domain_dataset}).
\begin{table}[ht]
\resizebox{\columnwidth}{!}{%
\begin{tabular}{l|l}
\hline
\textbf{Domain} & \textbf{Datasets} \\ \hline
Medical & EMEA-V3 \citep{opus} \\ \hline
Legal & MultiEURLEX \citep{multieurlex} \\ \hline
IT & \begin{tabular}[c]{@{}l@{}}Ubuntu, KDE4, GNOME, \\ PHP, OpenOffice \citep{opus} \end{tabular} \\ \hline
Movie Subtitles & OpenSubtitles \citep{subtitles} \\ \hline
TED Talks & TED2020 \citep{ted} \\ \hline
General & \begin{tabular}[c]{@{}l@{}}Wikipedia \citep{wiki} \\ NTREX-128 \citep{ntrex} \end{tabular} \\ \hline
\end{tabular}%
}
\caption{Datasets used for each domain.}
\label{tab:domain_dataset}
\end{table}

For fine-tuning the model, we use English and Spanish as the source and target languages, respectively, across all six domains. When evaluating the fine-tuned models, English remains the source language, while the target languages are chosen based on their linguistic relatedness to Spanish, the pivot language used for fine-tuning. The target languages for evaluation are Portuguese, Italian, French, Czech, Polish, and Greek. By selecting English as the source language for both fine-tuning and evaluation, we ensure consistency across results while potentially benefiting from shared linguistic properties between English and the target languages. The test data for each domain is parallel across all languages used in the experiments.

\begin{table*}[ht]
\resizebox{\textwidth}{!}{%
\begin{tabular}{l|rr|rr|rr|rr|rr|rr|rr}
\hline
\multirow{2}{*}{} &
  \multicolumn{2}{c|}{en→es} &
  \multicolumn{2}{c|}{en→pt} &
  \multicolumn{2}{c|}{en→it} &
  \multicolumn{2}{c|}{en→fr} &
  \multicolumn{2}{c|}{en→cs} &
  \multicolumn{2}{c|}{en→pl} &
  \multicolumn{2}{c}{en→el} \\ 
 &
  \multicolumn{1}{l}{BLEU} &
  \multicolumn{1}{l|}{COMET} &
  \multicolumn{1}{l}{BLEU} &
  \multicolumn{1}{l|}{COMET} &
  \multicolumn{1}{l}{BLEU} &
  \multicolumn{1}{l|}{COMET} &
  \multicolumn{1}{l}{BLEU} &
  \multicolumn{1}{l|}{COMET} &
  \multicolumn{1}{l}{BLEU} &
  \multicolumn{1}{l|}{COMET} &
  \multicolumn{1}{l}{BLEU} &
  \multicolumn{1}{l|}{COMET} &
  \multicolumn{1}{l}{BLEU} &
  \multicolumn{1}{l}{COMET} \\ \hline
med\_base &
  \multicolumn{1}{r}{36} &
  0.8451 &
  \multicolumn{1}{r}{31.7} &
  0.8527 &
  \multicolumn{1}{r}{29.7} &
  0.8481 &
  \multicolumn{1}{r}{28.6} &
  0.8179 &
  \multicolumn{1}{r}{\textbf{22.6}} &
  0.8582 &
  \multicolumn{1}{r}{18} &
  0.8234 &
  \multicolumn{1}{r}{\textbf{29.1}} &
  0.8595 \\
med\_ft &
  \multicolumn{1}{r}{\textbf{43.5}$\uparrow$} &
  \textbf{0.8663}$\uparrow$ &
  \multicolumn{1}{r}{\textbf{34.4}$\uparrow$} &
  \textbf{0.8645}$\uparrow$ &
  \multicolumn{1}{r}{\textbf{30.6}$\uparrow$} &
  \textbf{0.8563}$\uparrow$ &
  \multicolumn{1}{r}{\textbf{29.2}$\uparrow$} &
  \textbf{0.8293}$\uparrow$ &
  \multicolumn{1}{r}{22.5} &
  \textbf{0.8659}$\uparrow$ &
  \multicolumn{1}{r}{\textbf{18.1}} &
  \textbf{0.8342}$\uparrow$ &
  \multicolumn{1}{r}{28.8} &
  \textbf{0.8644}$\uparrow$ \\ \hline
leg\_base &
  \multicolumn{1}{r}{44.1} &
  0.8494 &
  \multicolumn{1}{r}{42.9} &
  0.8653 &
  \multicolumn{1}{r}{36.9} &
  0.8721 &
  \multicolumn{1}{r}{41.8} &
  0.8533 &
  \multicolumn{1}{r}{30.6} &
  0.8905 &
  \multicolumn{1}{r}{33.2} &
  0.8796 &
  \multicolumn{1}{r}{40.7} &
  0.8972 \\
leg\_ft &
  \multicolumn{1}{r}{\textbf{49.6}$\uparrow$} &
  \textbf{0.8714}$\uparrow$ &
  \multicolumn{1}{r}{\textbf{43.3}} &
  \textbf{0.8774}$\uparrow$ &
  \multicolumn{1}{r}{\textbf{38.3}$\uparrow$} &
  \textbf{0.8827}$\uparrow$ &
  \multicolumn{1}{r}{\textbf{43.7}$\uparrow$} &
  \textbf{0.8694}$\uparrow$ &
  \multicolumn{1}{r}{\textbf{33.4}$\uparrow$} &
  \textbf{0.9038}$\uparrow$ &
  \multicolumn{1}{r}{\textbf{34.2}$\uparrow$} &
  \textbf{0.8906}$\uparrow$ &
  \multicolumn{1}{r}{\textbf{43.9}$\uparrow$} &
  \textbf{0.9059}$\uparrow$ \\ \hline
it\_base &
  \multicolumn{1}{r}{34.2} &
  0.7954 &
  \multicolumn{1}{r}{26.6} &
  0.79 &
  \multicolumn{1}{r}{29.2} &
  0.7964 &
  \multicolumn{1}{r}{25.2} &
  0.7386 &
  \multicolumn{1}{r}{19.1} &
  0.7872 &
  \multicolumn{1}{r}{21.2} &
  0.7818 &
  \multicolumn{1}{r}{27.1} &
  0.7972 \\
it\_ft &
  \multicolumn{1}{r}{\textbf{44.4}$\uparrow$} &
  \textbf{0.8344}$\uparrow$ &
  \multicolumn{1}{r}{\textbf{29.5}$\uparrow$} &
  \textbf{0.8104}$\uparrow$ &
  \multicolumn{1}{r}{\textbf{33.4}$\uparrow$} &
  \textbf{0.8151}$\uparrow$ &
  \multicolumn{1}{r}{\textbf{29.5}$\uparrow$} &
  \textbf{0.7619}$\uparrow$ &
  \multicolumn{1}{r}{\textbf{19.7}} &
  \textbf{0.8147}$\uparrow$ &
  \multicolumn{1}{r}{\textbf{22.5}$\uparrow$} &
  \textbf{0.8078}$\uparrow$ &
  \multicolumn{1}{r}{\textbf{28.5}$\uparrow$} &
  \textbf{0.816}$\uparrow$ \\ \hline\hline
sub\_base &
  \multicolumn{1}{r}{22.8} &
  0.7548 &
  \multicolumn{1}{r}{20} &
  0.7678 &
  \multicolumn{1}{r}{17.8} &
  0.7446 &
  \multicolumn{1}{r}{16.1} &
  0.6988 &
  \multicolumn{1}{r}{15} &
  0.7594 &
  \multicolumn{1}{r}{14.6} &
  0.7472 &
  \multicolumn{1}{r}{\textbf{15.7}} &
  \textbf{0.7768} \\
sub\_ft &
  \multicolumn{1}{r}{\textbf{24.7}$\uparrow$} &
  \textbf{0.7627}$\uparrow$ &
  \multicolumn{1}{r}{\textbf{21.2}$\uparrow$} &
  \textbf{0.7741}$\uparrow$ &
  \multicolumn{1}{r}{\textbf{18.8}$\uparrow$} &
  \textbf{0.7532}$\uparrow$ &
  \multicolumn{1}{r}{\textbf{17.5}$\uparrow$} &
  \textbf{0.7073}$\uparrow$ &
  \multicolumn{1}{r}{\textbf{15.3}} &
  \textbf{0.766}$\uparrow$ &
  \multicolumn{1}{r}{\textbf{15.1}} &
  \textbf{0.7566}$\uparrow$ &
  \multicolumn{1}{r}{15.6} &
  0.7725 \\ \hline
ted\_base &
  \multicolumn{1}{r}{35.9} &
  0.8189 &
  \multicolumn{1}{r}{31.1} &
  0.8238 &
  \multicolumn{1}{r}{29.3} &
  0.814 &
  \multicolumn{1}{r}{33.7} &
  0.7926 &
  \multicolumn{1}{r}{21.8} &
  0.8082 &
  \multicolumn{1}{r}{16.2} &
  0.7857 &
  \multicolumn{1}{r}{29} &
  0.8439 \\
ted\_ft &
  \multicolumn{1}{r}{\textbf{37.8}$\uparrow$} &
  \textbf{0.8262}$\uparrow$ &
  \multicolumn{1}{r}{\textbf{31.9}$\uparrow$} &
  \textbf{0.8303}$\uparrow$ &
  \multicolumn{1}{r}{\textbf{29.6}} &
  \textbf{0.8199}$\uparrow$ &
  \multicolumn{1}{r}{\textbf{35}$\uparrow$} &
  \textbf{0.7955} &
  \multicolumn{1}{r}{\textbf{21.9}} &
  \textbf{0.8194}$\uparrow$ &
  \multicolumn{1}{r}{\textbf{16.6}$\uparrow$} &
  \textbf{0.7934}$\uparrow$ &
  \multicolumn{1}{r}{\textbf{29.2}} &
  \textbf{0.8475}$\uparrow$ \\ \hline\hline
gen\_base &
  \multicolumn{1}{r}{\textbf{32.7}} &
  0.786 &
  \multicolumn{1}{r}{30.5} &
  0.8032 &
  \multicolumn{1}{r}{\textbf{31.5}} &
  \textbf{0.8004} &
  \multicolumn{1}{r}{\textbf{26.1}} &
  \textbf{0.7534} &
  \multicolumn{1}{r}{\textbf{25.1}} &
  0.7831 &
  \multicolumn{1}{r}{\textbf{18.4}} &
  \textbf{0.763} &
  \multicolumn{1}{r}{\textbf{26.2}} &
  \textbf{0.8138} \\
gen\_ft &
  \multicolumn{1}{r}{32.5} &
  \textbf{0.7868} &
  \multicolumn{1}{r}{\textbf{30.8}} &
  \textbf{0.8043} &
  \multicolumn{1}{r}{31.3} &
  0.8002 &
  \multicolumn{1}{r}{25.1$\downarrow$} &
  0.753 &
  \multicolumn{1}{r}{24.7} &
  \textbf{0.7874} &
  \multicolumn{1}{r}{18.1} &
  0.7604 &
  \multicolumn{1}{r}{25.4$\downarrow$} &
  0.8065$\downarrow$ \\ \hline
\end{tabular}%
}
\caption{Main results, comparing the performance of the baseline (base) and fine-tuned (ft) models using BLEU and COMET scores across six domains: medical (med), legal (leg), IT (it), movie subtitles (sub), TED talks (ted), and a general domain (gen). \textbf{Bold} values indicate the higher score between the baseline and fine-tuned models for each metric, domain, and language pair. Arrows indicate significantly worse ($\downarrow$) and better ($\uparrow$) performance compared to the baseline according to each metric, domain, and language pair (p-value < 0.05).}
\label{tab:main_results}
\end{table*}

\subsection{Data Preprocessing}
To prepare the datasets for fine-tuning and evaluation, we apply a standardized preprocessing pipeline. For the OPUS-sourced datasets\footnote{\url{https://opus.nlpl.eu/}} in the medical, IT, movie subtitles, and TED talks domains, we download the English-to-Spanish parallel data as well as the English-to-target-language data for each of the six target languages. The Wikipedia dataset for English-to-Spanish is also obtained from OPUS to represent the general domain training and validation sets. For the multilingual legal domain data from MultiEURLEX,\footnote{\url{https://github.com/nlpaueb/multi-eurlex}} as well as the NTREX-128 test set\footnote{\url{https://github.com/MicrosoftTranslator/NTREX}} representing the general domain, we retrieve the corpora directly from their respective repositories. As the OPUS datasets are originally aligned at the sentence level, we first clean and filter the data using a consistent methodology across all domains. This involves removing sentences with token lengths outside the range of 3 to 100, irregular punctuation, duplicates, and sentences exhibiting a similarity of 60\% or higher. Any source or target sentences that do not meet these criteria are discarded from the parallel data in each domain to mitigate potential noise. Next, we preprocess the data to extract a set of 1,000 parallel sentences as the test data for all eight languages (English and the seven target languages) in each domain. 

These parallel test sentences are then removed from the English-Spanish parallel dataset, with the remaining data split into a validation set of 1,000 sentences and a training set of 150,000 sentences for each domain. For the NTREX-128 test set, we do not apply any data cleaning and simply extract the first 1,000 sentences for each of the eight languages to create the general domain test sets. The MultiEURLEX corpus, being document-aligned, requires aligning the data at the sentence level for the training, validation, and test splits provided in the corpus before the cleaning steps applied to the OPUS datasets.
\subsection{Model}
\label{model}
The baseline model employed in our experiments is the M2M-100 (many-to-many for 100 languages) multilingual NMT model \citep{multi2}. M2M-100 is a sequence-to-sequence Transformer model capable of translating directly between any pair of its supported 100 languages without relying on English as an intermediary. The original M2M-100 model was trained on a diverse parallel corpus spanning 100 languages, curated through a novel data mining approach called the ``bridge language family mining strategy'' \citep{multi2} and mined from the Common Crawl corpus.\footnote{\url{https://commoncrawl.org/}} While all languages used in our experiments, except for Italian, are considered bridge languages in the M2M-100 model, the lack of information on the exact data sizes mined for each language limits our ability to comprehensively analyze how the amount of pre-training data affects the language-specific performance of the model. For our experiments, we utilize the \verb+m2m100\_418M+ variant\footnote{\url{https://huggingface.co/facebook/m2m100_418M/tree/main}} with 418 million parameters to meet the hardware limitations of our project. 
\subsection{Implementation}
\label{implementation}
The implementation of the M2M-100 model is based on the Transformers library from Hugging Face.\footnote{\url{https://huggingface.co/docs/transformers/en/index}} We fine-tune the model on the English-Spanish parallel dataset for each of the six domains separately, using the same set of configurations, including a learning rate of 1e-7, a batch size of 10, dropout of 0.1, weight decay of 0.0, label smoothing of 0.2, AdamW optimizer with betas of 0.9 and 0.98, a maximum input/output length of 128 tokens, mixed precision (FP16) training, and a maximum of 60,000 training steps, with epoch-level validation. Due to resource limitations, we do not train the models until convergence. We freeze the embedding layers of the encoder to prevent catastrophic forgetting of the pre-trained representations. The models are trained on a single NVIDIA-T4 GPU (each for around seven hours), and the best checkpoint is saved for inference. For inference, we load each fine-tuned M2M-100 model and its corresponding tokenizer and generate the translated output using beam search decoding with a beam size of 4. The baselines involve evaluating the initial \verb+m2m100\_418M+ checkpoint on all target languages in each domain separately, using the same inference configurations as for the fine-tuned models. All experiments are conducted in the Google Colab environment.\footnote{\url{https://colab.research.google.com/}}\footnote{All models and data used in the experiments will be made public in the final version.}

We include BLEU \citep{bleu} as one of our evaluation metrics due to its widespread adoption within the MT research. We rely on sacreBLEU \citep{bleu_clarity} (\verb+13a+ tokenizer) and its implementation of paired bootstrap resampling \citep{sig_test} with 300 resampling trials and a p-value threshold of 0.05. We also use COMET \citep{comet}, for which we employ \verb+comet-compare+ (bootstrap resampling and a paired t-test, 300 resamples and a p-value of 0.05). 
\section{Results}
\label{results}
First, we examine the effectiveness of fine-tuning a massively multilingual pre-trained model, M2M-100, on domain-specific data from an English-Spanish language pair and evaluate its performance across various target languages within the same domain.

Table \ref{tab:main_results} presents the main results, comparing the pre-trained baseline model's performance against the fine-tuned models across six domains. Across the specialized medical, legal, and IT domains, the fine-tuned models consistently outperform the baseline model, achieving higher BLEU and COMET scores for all language pairs (see Appendix \ref{appd:full_results} for CometKiwi scores), with few exceptions depending on the evaluation metric. This improvement demonstrates the effectiveness of domain adaptation through fine-tuning, enabling the model to capture domain-specific knowledge and vocabulary more effectively within well-defined, specialized domains. 

However, the degree of improvement varies across domains and target languages. For the medical domain, while fine-tuning significantly improves the results for the English-Spanish language pair used for fine-tuning, the improvement becomes less pronounced as the target language diverges linguistically from the pivot language (Spanish). This trend is less evident in the legal and IT domains, where substantial improvements are observed across almost all language pairs, even for those linguistically distant from Spanish. In contrast, for the mixed domains of movie subtitles and TED talks, which lack a specific topical focus but exhibit distinct genres, the improvements from fine-tuning are more modest, although still consistent across most language pairs. Notably, the improvements are more consistent for the TED talks domain than the movie subtitles domain. Moreover, even the performance for the English-Spanish language pair used for fine-tuning shows only marginal improvements compared to the specialized domains. In the general domain, the fine-tuned model exhibits mixed performance, with statistically insignificant improvements in some language pairs and slight decreases in others compared to the baseline model. This could be attributed to the composition of the pre-training data, which contains substantial general-domain content, potentially limiting the benefits of fine-tuning on a specific language pair, as the baseline model is already well-trained on such data. 
\begin{table*}[ht]
\resizebox{\textwidth}{!}{%
\begin{tabular}{l|rr|rr|rr|rr|rr|rr|rr}
\hline
 &
  \multicolumn{2}{c|}{en→es} &
  \multicolumn{2}{c|}{en→pt} &
  \multicolumn{2}{c|}{en→it} &
  \multicolumn{2}{c|}{en→fr} &
  \multicolumn{2}{c|}{en→cs} &
  \multicolumn{2}{c|}{en→pl} &
  \multicolumn{2}{c}{en→el} \\
\multirow{-2}{*}{} &
  \multicolumn{1}{l}{BLEU} &
  \multicolumn{1}{l|}{COMET} &
  \multicolumn{1}{l}{BLEU} &
  \multicolumn{1}{l|}{COMET} &
  \multicolumn{1}{l}{BLEU} &
  \multicolumn{1}{l|}{COMET} &
  \multicolumn{1}{l}{BLEU} &
  \multicolumn{1}{l|}{COMET} &
  \multicolumn{1}{l}{BLEU} &
  \multicolumn{1}{l|}{COMET} &
  \multicolumn{1}{l}{BLEU} &
  \multicolumn{1}{l|}{COMET} &
  \multicolumn{1}{l}{BLEU} &
  \multicolumn{1}{l}{COMET} \\ \hline
med\_base &
  \multicolumn{1}{r}{\textbf{36}} &
  \textbf{0.8451} &
  \multicolumn{1}{r}{\textbf{31.7}} &
  0.8527 &
  \multicolumn{1}{r}{\textbf{29.7}} &
  \textbf{0.8481} &
  \multicolumn{1}{r}{\textbf{28.6}} &
  \textbf{0.8179} &
  \multicolumn{1}{r}{\textbf{22.6}} &
  0.8582 &
  \multicolumn{1}{r}{\textbf{18}} &
  \textbf{0.8234} &
  \multicolumn{1}{r}{\textbf{29.1}} &
  \textbf{0.8595} \\
sub2med &
  \multicolumn{1}{r}{35.3$\downarrow$} &
  0.8416$\downarrow$ &
  \multicolumn{1}{r}{30.4$\downarrow$} &
  0.8473$\downarrow$ &
  \multicolumn{1}{r}{28.6$\downarrow$} &
  0.8444$\downarrow$ &
  \multicolumn{1}{r}{27.9$\downarrow$} &
  0.8133$\downarrow$ &
  \multicolumn{1}{r}{21.6$\downarrow$} &
  0.8506$\downarrow$ &
  \multicolumn{1}{r}{16.9$\downarrow$} &
  0.815$\downarrow$ &
  \multicolumn{1}{r}{27.4$\downarrow$} &
  0.8491$\downarrow$ \\
ted2med &
  \multicolumn{1}{r}{35.7} &
  0.8444 &
  \multicolumn{1}{r}{31.4} &
  \textbf{0.8528} &
  \multicolumn{1}{r}{29.4} &
  0.8477 &
  \multicolumn{1}{r}{28$\downarrow$} &
  0.8155 &
  \multicolumn{1}{r}{22.1} &
  \textbf{0.8595} &
  \multicolumn{1}{r}{17.4$\downarrow$} &
  0.8217 &
  \multicolumn{1}{r}{28.4$\downarrow$} &
  0.8585 \\
\rowcolor[HTML]{E2E0E0} 
{\color[HTML]{504E4E} med\_ft} &
  \multicolumn{1}{r}{\cellcolor[HTML]{E2E0E0}{\color[HTML]{504E4E} \textbf{43.5}$\uparrow$}} &
  {\color[HTML]{504E4E} \textbf{0.8663}$\uparrow$} &
  \multicolumn{1}{r}{\cellcolor[HTML]{E2E0E0}{\color[HTML]{504E4E} \textbf{34.4}$\uparrow$}} &
  {\color[HTML]{504E4E} \textbf{0.8645}$\uparrow$} &
  \multicolumn{1}{r}{\cellcolor[HTML]{E2E0E0}{\color[HTML]{504E4E} \textbf{30.6}$\uparrow$}} &
  {\color[HTML]{504E4E} \textbf{0.8563}$\uparrow$} &
  \multicolumn{1}{r}{\cellcolor[HTML]{E2E0E0}{\color[HTML]{504E4E} \textbf{29.2}$\uparrow$}} &
  {\color[HTML]{504E4E} \textbf{0.8293}$\uparrow$} &
  \multicolumn{1}{r}{\cellcolor[HTML]{E2E0E0}{\color[HTML]{504E4E} 22.5}} &
  {\color[HTML]{504E4E} \textbf{0.8659}$\uparrow$} &
  \multicolumn{1}{r}{\cellcolor[HTML]{E2E0E0}{\color[HTML]{504E4E} \textbf{18.1}}} &
  {\color[HTML]{504E4E} \textbf{0.8342}$\uparrow$} &
  \multicolumn{1}{r}{\cellcolor[HTML]{E2E0E0}{\color[HTML]{504E4E} 28.8}} &
  {\color[HTML]{504E4E} \textbf{0.8644}$\uparrow$} \\ \hline\hline
leg\_base &
  \multicolumn{1}{r}{\textbf{44.1}} &
  0.8494 &
  \multicolumn{1}{r}{\textbf{42.9}} &
  0.8653 &
  \multicolumn{1}{r}{\textbf{36.9}} &
  \textbf{0.8721} &
  \multicolumn{1}{r}{41.8} &
  0.8533 &
  \multicolumn{1}{r}{30.6} &
  0.8905 &
  \multicolumn{1}{r}{\textbf{33.2}} &
  0.8796 &
  \multicolumn{1}{r}{40.7} &
  0.8972 \\
sub2leg &
  \multicolumn{1}{r}{41.6$\downarrow$} &
  0.8424$\downarrow$ &
  \multicolumn{1}{r}{40.3$\downarrow$} &
  0.8614$\downarrow$ &
  \multicolumn{1}{r}{35.1$\downarrow$} &
  0.8645$\downarrow$ &
  \multicolumn{1}{r}{40.2$\downarrow$} &
  0.8514 &
  \multicolumn{1}{r}{29.6$\downarrow$} &
  0.8866$\downarrow$ &
  \multicolumn{1}{r}{30.7$\downarrow$} &
  0.8729$\downarrow$ &
  \multicolumn{1}{r}{38.4$\downarrow$} &
  0.8919$\downarrow$ \\
ted2leg &
  \multicolumn{1}{r}{43.4$\downarrow$} &
  \textbf{0.8503} &
  \multicolumn{1}{r}{42.1$\downarrow$} &
  \textbf{0.8671} &
  \multicolumn{1}{r}{36.6} &
  0.8717 &
  \multicolumn{1}{r}{\textbf{42}} &
  \textbf{0.8589$\uparrow$} &
  \multicolumn{1}{r}{\textbf{31.3}$\uparrow$} &
  \textbf{0.8923} &
  \multicolumn{1}{r}{32.8} &
  \textbf{0.8823} &
  \multicolumn{1}{r}{\textbf{41.3}$\uparrow$} &
  \textbf{0.899} \\
\rowcolor[HTML]{E2E0E0} 
{\color[HTML]{504E4E} leg\_ft} &
  \multicolumn{1}{r}{\cellcolor[HTML]{E2E0E0}{\color[HTML]{504E4E} \textbf{49.6}$\uparrow$}} &
  {\color[HTML]{504E4E} \textbf{0.8714}$\uparrow$} &
  \multicolumn{1}{r}{\cellcolor[HTML]{E2E0E0}{\color[HTML]{504E4E} \textbf{43.3}}} &
  {\color[HTML]{504E4E} \textbf{0.8774}$\uparrow$} &
  \multicolumn{1}{r}{\cellcolor[HTML]{E2E0E0}{\color[HTML]{504E4E} \textbf{38.3}$\uparrow$}} &
  {\color[HTML]{504E4E} \textbf{0.8827}$\uparrow$} &
  \multicolumn{1}{r}{\cellcolor[HTML]{E2E0E0}{\color[HTML]{504E4E} \textbf{43.7}$\uparrow$}} &
  {\color[HTML]{504E4E} \textbf{0.8694}$\uparrow$} &
  \multicolumn{1}{r}{\cellcolor[HTML]{E2E0E0}{\color[HTML]{504E4E} \textbf{33.4}$\uparrow$}} &
  {\color[HTML]{504E4E} \textbf{0.9038}$\uparrow$} &
  \multicolumn{1}{r}{\cellcolor[HTML]{E2E0E0}{\color[HTML]{504E4E} \textbf{34.2}$\uparrow$}} &
  {\color[HTML]{504E4E} \textbf{0.8906}$\uparrow$} &
  \multicolumn{1}{r}{\cellcolor[HTML]{E2E0E0}{\color[HTML]{504E4E} \textbf{43.9}$\uparrow$}} &
  {\color[HTML]{504E4E} \textbf{0.9059}$\uparrow$} \\ \hline\hline
it\_base &
  \multicolumn{1}{r}{34.2} &
  0.7954 &
  \multicolumn{1}{r}{26.6} &
  0.79 &
  \multicolumn{1}{r}{29.2} &
  0.7964 &
  \multicolumn{1}{r}{25.2} &
  0.7386 &
  \multicolumn{1}{r}{19.1} &
  0.7872 &
  \multicolumn{1}{r}{21.2} &
  0.7818 &
  \multicolumn{1}{r}{27.1} &
  0.7972 \\
sub2it &
  \multicolumn{1}{r}{\textbf{34.2}} &
  0.7929 &
  \multicolumn{1}{r}{\textbf{26.6}} &
  0.7866 &
  \multicolumn{1}{r}{\textbf{29.9}} &
  0.7962 &
  \multicolumn{1}{r}{24.7} &
  0.7335$\downarrow$ &
  \multicolumn{1}{r}{18$\downarrow$} &
  0.7829 &
  \multicolumn{1}{r}{20.3$\downarrow$} &
  0.7793 &
  \multicolumn{1}{r}{25.7$\downarrow$} &
  0.7826$\downarrow$ \\
ted2it &
  \multicolumn{1}{r}{\textbf{35.9}$\uparrow$} &
  \textbf{0.8032}$\uparrow$ &
  \multicolumn{1}{r}{\textbf{28.2}$\uparrow$} &
  \textbf{0.7968}$\uparrow$ &
  \multicolumn{1}{r}{\textbf{30.8}$\uparrow$} &
  \textbf{0.8023}$\uparrow$ &
  \multicolumn{1}{r}{\textbf{26.5}$\uparrow$} &
  \textbf{0.7451}$\uparrow$ &
  \multicolumn{1}{r}{\textbf{19.7}} &
  \textbf{0.7942}$\uparrow$ &
  \multicolumn{1}{r}{\textbf{21.7}} &
  \textbf{0.7898}$\uparrow$ &
  \multicolumn{1}{r}{\textbf{28.3}$\uparrow$} &
  \textbf{0.8009} \\ 
\rowcolor[HTML]{E2E0E0} 
{\color[HTML]{504E4E} it\_ft} &
  \multicolumn{1}{r}{\cellcolor[HTML]{E2E0E0}{\color[HTML]{504E4E} \textbf{44.4}$\uparrow$}} &
  {\color[HTML]{504E4E} \textbf{0.8344}$\uparrow$} &
  \multicolumn{1}{r}{\cellcolor[HTML]{E2E0E0}{\color[HTML]{504E4E} \textbf{29.5}$\uparrow$}} &
  {\color[HTML]{504E4E} \textbf{0.8104}$\uparrow$} &
  \multicolumn{1}{r}{\cellcolor[HTML]{E2E0E0}{\color[HTML]{504E4E} \textbf{33.4}$\uparrow$}} &
  {\color[HTML]{504E4E} \textbf{0.8151}$\uparrow$} &
  \multicolumn{1}{r}{\cellcolor[HTML]{E2E0E0}{\color[HTML]{504E4E} \textbf{29.5}$\uparrow$}} &
  {\color[HTML]{504E4E} \textbf{0.7619}$\uparrow$} &
  \multicolumn{1}{r}{\cellcolor[HTML]{E2E0E0}{\color[HTML]{504E4E} \textbf{19.7}}} &
  {\color[HTML]{504E4E} \textbf{0.8147}$\uparrow$} &
  \multicolumn{1}{r}{\cellcolor[HTML]{E2E0E0}{\color[HTML]{504E4E} \textbf{22.5}$\uparrow$}} &
  {\color[HTML]{504E4E} \textbf{0.8078}$\uparrow$} &
  \multicolumn{1}{r}{\cellcolor[HTML]{E2E0E0}{\color[HTML]{504E4E} \textbf{28.5}$\uparrow$}} &
  {\color[HTML]{504E4E} \textbf{0.816}$\uparrow$} \\ \hline
\end{tabular}%
}
\caption{Results for zero-shot cross-lingual cross-domain transfer, where models fine-tuned on movie subtitles (sub) or TED talks (ted) are evaluated on medical (med), legal (leg), and IT (it) domains using BLEU and COMET scores. Scores compare cross-domain transfer models (sub2med, ted2med, etc.) against baseline (base) and models fine-tuned on the target domain (med\_ft, leg\_ft, it\_ft). \textbf{Bold} values indicate higher score between the baseline and fine-tuned models for each metric, domain, and language pair. \colorbox{lightgray}{Gray} rows show expected higher scores (also in \textbf{bold}) for direct fine-tuning. Arrows indicate significantly worse ($\downarrow$) and better ($\uparrow$) performance compared to the baseline according to each metric, domain, and language pair (p-value < 0.05).}
\label{tab:domain_transfer}
\end{table*}
Turning to the subject of domain transferability, Table \ref{tab:domain_transfer} illustrates the results of zero-shot cross-lingual cross-domain transfer, where the model is fine-tuned on movie subtitles or TED talks mixed domains and evaluated on the specialized medical, legal, and IT domains across all language pairs. The results exhibit mixed performance, depending on the domain and language pair, with improvements generally more pronounced, while not statistically significant, when evaluated using COMET than BLEU. Notably, fine-tuning on TED talks leads to more effective cross-domain transfer compared to movie subtitles. Regarding the target specialized domains, cross-domain transfer appears more effective for the IT domain than legal or medical domains. This observation suggests certain domains may be more responsive to cross-domain transfer due to their inherent characteristics. 
\section{Discussion}
\subsection{Enhancing Domain-Specific Translation Quality}
The main results (see Table \ref{tab:main_results}) imply that the domain-specific quality of MT output for one language pair can indeed be enhanced by fine-tuning the model on domain-relevant data from another language pair. The fine-tuned models consistently outperform the baseline across the specialized medical, legal, and IT domains, achieving higher BLEU and COMET scores for almost all language pairs. This improvement demonstrates the effectiveness of domain adaptation through fine-tuning, enabling the model to capture domain-specific knowledge and vocabulary more effectively within well-defined domains. While the degree of improvement varies across domains and target languages, the consistent gains observed underscore the potential of zero-shot cross-lingual domain adaptation. By utilizing domain-specific data from a single language pair, the model's performance can be enhanced for multiple target languages within the same domain, even for those linguistically distant from the fine-tuning language pair.
\subsection{Factors Influencing Domain Transferability}
The results presented in Tables \ref{tab:main_results} and \ref{tab:domain_transfer} reveal varying levels of improvement in zero-shot cross-lingual domain adaptation across specialized domains, as well as differences in how responsive certain domains are to cross-domain transfer. To investigate the factors behind these observations, we take a deeper look into the characteristics of each domain. One key factor that emerges is the linguistic complexity of the domains, as evidenced by the average sentence length and vocabulary size shown in Table \ref{tab:sent_len}. The movie subtitles domain has approximately twice the shorter average sentence length and a 20\% smaller vocabulary size in the training set compared to the TED talks domain, which could explain why fine-tuning on the more complex TED talks domain results in better generalization to specialized domains.
\begin{table*}[ht]
\centering
\resizebox{\textwidth}{!}{%
\begin{tabular}{l|rr|rr|rr}
\hline
Domain & \multicolumn{2}{c|}{Train} & \multicolumn{2}{c|}{Valid} & \multicolumn{2}{c}{Test} \\
 & \multicolumn{1}{l}{Avg Sent Len} & \multicolumn{1}{l|}{Vocab Size} & \multicolumn{1}{l}{Avg Sent Len} & \multicolumn{1}{l|}{Vocab Size} & \multicolumn{1}{l}{Avg Sent Len} & \multicolumn{1}{l}{Vocab Size} \\ \hline
Medical & 19.22 & 23,824 & 19.20 & 3,462 & 16.93 & 2,966 \\ \hline
Legal & 36.33 & 33,657 & 33.45 & 2,971 & 28.57 & 2,975 \\ \hline
IT & 12.07 & 25,111 & 12.54 & 2,428 & 9.41 & 1,830 \\ \hline
Movie subtitles & 9.76 & 34,310 & 10.35 & 2,193 & 10.34 & 2,117 \\ \hline
TED talks & 18.10 & 42,513 & 18.75 & 3,185 & 17.43 & 3,421 \\ \hline
\end{tabular}%
}
\caption{Average sentence length (in words) and vocabulary size (number of unique words) for the training, validation, and test sets across specialized and mixed domains.}
\label{tab:sent_len}
\end{table*}
Among the specialized domains, the IT domain stands out with the smallest average sentence length and vocabulary size, potentially making it easier for models to generalize and adapt to this domain. In contrast, the legal domain appears to be the most linguistically complex in terms of sentence length and vocabulary size. However, certain unique aspects of legal language may paradoxically facilitate transfer to this domain. The unique language style prevalent in legal texts, featuring archaic vocabulary, repetitive syntax patterns, and convoluted sentences can, while increasing overall complexity, also aid models in adapting to the legal domain's language use. Additionally, the substantially longer average sentence length in the legal domain (around three times longer than the IT domain and two times longer than the medical domain) means that when training for a fixed number of epochs, the model was exposed to more training data in terms of the total number of words. This increased exposure to legal language likely contributed to better adaptation to this domain. 

These findings demonstrate that domain transferability is not solely determined by a domain's topical specificity or genre but is also influenced by its linguistic properties and inherent characteristics. While domains exhibiting greater linguistic complexity in terms of sentence length and vocabulary size tend to be more challenging for effective cross-lingual adaptation and cross-domain transfer, certain linguistic features can actually facilitate transfer. Consequently, in addition to the domain's topical and genre specialization, a nuanced understanding of a domain's inherent linguistic characteristics, as well as the amount of exposure to a domain's language during training, is required for optimizing cross-lingual adaptation and cross-domain transfer. Fundamental properties like vocabulary size, sentence complexity, and consistent language styles emerge as key factors influencing transferability potential across domains.
\subsection{Influence of Language-Specific and Domain-Specific Factors}
The results also show that both language-specific and domain-specific factors have influence on the task, with language influences being more prominent in the main results, while domain-specific factors play a crucial role in determining cross-domain transfer effectiveness. The main results (see Table \ref{tab:main_results}) highlight language influences, with larger gains observed for the fine-tuning language pair (English-Spanish) and less pronounced improvements as the linguistic distance from the pivot language increases. This trend shows the role of language-specific factors and linguistic proximity in the effectiveness of cross-lingual domain adaptation. However, domain-specific factors also play an essential role, as evident from the varied performance across specialized domains discussed in the previous section.

The findings suggest that while fine-tuning the model on closely related language pairs is advantageous, inherent domain characteristics ultimately determine the limits of both zero-shot cross-lingual domain adaptation and cross-domain transfer. Additionally, as shown in Table \ref{tab:trans_examples}, the specificity of domain vocabularies in specialized domains, where terminologies, though highly specialized, are consistent across languages, facilitates language transfer even for linguistically distant languages. This characteristic can be attributed to the presence of loanwords, which are fully or partly assimilated from one language into another, or terms that often remain untranslated. Furthermore, the influence of the pre-training data cannot be overlooked. Seven out of the eight languages involved in the experiments were considered bridge languages in the M2M-100 model, meaning the model was trained on comparatively larger amounts of data in these languages, capturing general language knowledge (the only exception is Italian, which, while not a bridge language in M2M-100, is still considered a relatively high-resource language). This pre-existing language knowledge likely contributes to the observed language transfer capabilities.
\begin{table*}[ht]
\centering
\resizebox{\textwidth}{!}{%
\begin{tabular}{p{1.2\textwidth}}
\hline
\textbf{Medical Domain} \\ \hline
\textbf{Source Text:} Like all medicines, \textcolor{red}{KOGENATE Bayer} 1000 IU can cause side effects, although not everybody gets them. \\
\textbf{Italian Reference:} Come tutti i medicinali, \textcolor{blue}{KOGENATE Bayer} 1000 UI può causare effetti indesiderati sebbene non tutte le persone li manifestino. \\
\textbf{Polish Reference:} Jak każdy lek, preparat \textcolor{blue}{KOGENATE Bayer} 1000 j. m. może powodować działania niepożądane, chociaż nie u każdego one wystąpią. \\ \hline
 \\
\textbf{Legal Domain} \\ \hline
\textbf{Source Text:} 5. As soon as it adopts a delegated act, the \textcolor{red}{Commission} shall notify it simultaneously to the \textcolor{red}{European Parliament} and to the Council. \\
\textbf{Italian Reference:} 5. Non appena adotta un atto delegato, la \textcolor{blue}{Commissione} ne dà contestualmente notifica al \textcolor{blue}{Parlamento europeo} e al Consiglio. \\
\textbf{Polish Reference:} 5. Niezwłocznie po przyjęciu aktu delegowanego \textcolor{blue}{Komisja} przekazuje go równocześnie \textcolor{blue}{Parlamentowi Europejskiemu} i Radzie. \\ \hline
 \\
\textbf{IT Domain} \\ \hline
\textbf{Source Text:} The \textcolor{red}{ISTIME() function} returns \textcolor{red}{True} if the \textcolor{red}{parameter} is a time value. Otherwise, it returns \textcolor{red}{False}. \\
\textbf{Italian Reference:} La \textcolor{blue}{funzione ISTIME()} restituisce \textcolor{blue}{True} se il \textcolor{blue}{parametro} è un' espressione di tempo. Altrimenti restituisce \textcolor{blue}{False}. \\
\textbf{Polish Reference:} \textcolor{blue}{Funkcja ISTIME()} zwraca \textcolor{blue}{True} jeśli \textcolor{blue}{parametr} ma wartość czasu, w przeciwnym wypadku \textcolor{blue}{False}. \\ \hline
\end{tabular}%
}
\caption{Examples of reference translations in the medical, legal, and IT domain test sets across English-to-Italian and English-to-Polish language pairs, highlighting the presence of loanwords and untranslated terms, marked in \textcolor{red}{red} in the source text and in \textcolor{blue}{blue} in the references.}
\label{tab:trans_examples}
\end{table*}
Therefore, while language-specific factors play a more prominent role in the main task of zero-shot cross-lingual domain adaptation, domain-specific factors are crucial determinants of cross-domain transfer effectiveness. These findings highlight the importance of considering both language and domain aspects when adapting NMT systems for specialized domains. The results also demonstrate that domains can exhibit distinct linguistic properties outside the notions of topic and genre, which significantly impact the effectiveness of cross-lingual adaptation and cross-domain transfer.

Consequently, a more nuanced understanding of a domain's inherent linguistic characteristics is crucial for optimizing these processes. By emphasizing the influence of domain-specific factors on transfer performance, this study highlights the importance of revisiting the traditional definitions of ``domain'' in MT. Current research often skips over this distinction, relying primarily on topical or genre-based domain classifications. However, the findings highlight the need for a more comprehensive characterization that accounts for linguistic complexities and domain-specific language patterns to develop effective strategies for tailoring NMT systems to diverse specialized domains.
\section{Conclusion and Future Work}
The study explored zero-shot cross-lingual domain adaptation for NMT to bridge the gap between large multilingual models and their limited specialized domain performance. Experiments across six domains showed consistent translation quality improvements for most target languages compared to the pre-trained baseline when fine-tuning a pivot language from the same domain. However, the degree of improvement varied based on the linguistic proximity between pivot and target languages, as well as the domain's linguistic complexity and data variety. The feasibility of zero-shot cross-lingual cross-domain transfer, using models fine-tuned on mixed domains for specialized domains, was also investigated. While achievable, effectiveness depended on the properties of pivot and target domains, with more consistent language domains being more responsive to cross-domain transfer. Future work can explore a broader range of specialized domains/languages and focus specifically on cross-domain transfer techniques. Also, including diverse language families will enable a better understanding of how language characteristics interact with domain transferability potential.
\section{Limitations}
The findings of this research are subject to several limitations, the first and primary one being the use of an existing pre-trained model (M2M-100) rather than pre-training a model specifically for the languages and domains included in the experiments due to resource constraints. Pre-training a model from scratch would have allowed for better control over the pre-training data, ensuring minimal overlap with the domain-specific data used for fine-tuning. Furthermore, the models are not fine-tuned until convergence, which potentially impacts the full realization of their capabilities. Additionally, the experiments focus on languages primarily from the Indo-European family, limiting the insights into the influence of linguistic relatedness and transferability potential across more diverse language pairs. Addressing these limitations is crucial for future research to foster a more comprehensive representation of the proposed approach. 

\section{Ethical Considerations}

In this paper, we used open-source datasets and models that were already published. Therefore, there are no ethical considerations for both our models and results. 
\bibliography{acl_latex}

\appendix
\section{Complete Evaluation Results}
\label{appd:full_results}
In this section, we report full results, illustrating the translation performance on all language pairs for all models, evaluated using BLEU, COMET, and CometKiwi.
\begin{table*}[ht]
\begin{minipage}[b]{0.5\textwidth}
\centering
\begin{tabular}{l|rrr}
\hline
\multirow{2}{*}{} & \multicolumn{3}{c}{en→es} \\
 & \multicolumn{1}{l}{BLEU} & \multicolumn{1}{l}{COMET} & \multicolumn{1}{l}{CometKiwi} \\ \hline
med\_base & 36 & 0.8451 & 0.8355 \\
sub2med & 35.3$\downarrow$ & 0.8416$\downarrow$ & 0.8307$\downarrow$ \\
ted2med & 35.7 & 0.8444 & 0.8358 \\
med\_ft & \textbf{43.5}$\uparrow$ & \textbf{0.8663}$\uparrow$ & \textbf{0.8481}$\uparrow$ \\ \hline
leg\_base & 44.1 & 0.8494 & 0.8514 \\
sub2leg & 41.6$\downarrow$ & 0.8424$\downarrow$ & 0.8448$\downarrow$ \\
ted2leg & 43.4$\downarrow$ & 0.8503 & 0.8617$\uparrow$ \\
leg\_ft & \textbf{49.6}$\uparrow$ & \textbf{0.8714}$\uparrow$ & \textbf{0.8682}$\uparrow$ \\ \hline
it\_base & 34.2 & 0.7954 & 0.7681 \\
sub2it & 34.2 & 0.7929 & 0.7656 \\
ted2it & 35.9$\uparrow$ & 0.8032$\uparrow$ & 0.782$\uparrow$ \\
it\_ft & \textbf{44.4}$\uparrow$ & \textbf{0.8344}$\uparrow$ & \textbf{0.7979}$\uparrow$ \\ \hline\hline
sub\_base & 22.8 & 0.7548 & 0.767 \\
sub\_ft & \textbf{24.7}$\uparrow$ & \textbf{0.7627}$\uparrow$ & \textbf{0.771} \\ \hline
ted\_base & 35.9 & 0.8189 & 0.8042 \\
ted\_ft & \textbf{37.8}$\uparrow$ & \textbf{0.8262}$\uparrow$ & \textbf{0.8076}$\uparrow$ \\ \hline\hline
gen\_base & \textbf{32.7} & 0.786 & 0.7723 \\
gen\_ft & 32.5 & \textbf{0.7868} & \textbf{0.7736} \\ \hline
\end{tabular}
\end{minipage}
\hfill
\begin{minipage}[b]{0.5\textwidth}
\centering
\begin{tabular}{l|rrr}
\hline
\multirow{2}{*}{} & \multicolumn{3}{c}{en→pt} \\
 & \multicolumn{1}{l}{BLEU} & \multicolumn{1}{l}{COMET} & \multicolumn{1}{l}{CometKiwi} \\ \hline
med\_base & 31.7 & 0.8527 & 0.8285 \\
sub2med & 30.4$\downarrow$ & 0.8473$\downarrow$ & 0.8228$\downarrow$ \\
ted2med & 31.4 & 0.8528 & 0.8319$\uparrow$ \\
med\_ft & \textbf{34.4}$\uparrow$ & \textbf{0.8645}$\uparrow$ & \textbf{0.8383}$\uparrow$ \\ \hline
leg\_base & 42.9 & 0.8653 & 0.8407 \\
sub2leg & 40.3$\downarrow$ & 0.8614$\downarrow$ & 0.8377 \\
ted2leg & 42.1$\downarrow$ & 0.8671 & 0.851$\uparrow$ \\
leg\_ft & \textbf{43.3} & \textbf{0.8774}$\uparrow$ & \textbf{0.8559}$\uparrow$ \\ \hline
it\_base & 26.6 & 0.79 & 0.7685 \\
sub2it & 26.6 & 0.7866 & 0.7637$\downarrow$ \\
ted2it & 28.2$\uparrow$ & 0.7968$\uparrow$ & 0.7788$\uparrow$ \\
it\_ft & \textbf{29.5}$\uparrow$ & \textbf{0.8104}$\uparrow$ & \textbf{0.7886}$\uparrow$ \\ \hline\hline
sub\_base & 20 & 0.7678 & 0.7694 \\
sub\_ft & \textbf{21.2}$\uparrow$ & \textbf{0.7741}$\uparrow$ & \textbf{0.7722} \\ \hline
ted\_base & 31.1 & 0.8238 & 0.7979 \\
ted\_ft & \textbf{31.9}$\uparrow$ & \textbf{0.8303}$\uparrow$ & \textbf{0.8032}$\uparrow$ \\ \hline\hline
gen\_base & 30.5 & 0.8032 & \textbf{0.7705} \\
gen\_ft & \textbf{30.8} & \textbf{0.8043} & \textbf{0.7705} \\ \hline
\end{tabular}
\end{minipage}
\caption{Translation performance on en→es and en→pt language pairs for all models, evaluated using BLEU, COMET, and CometKiwi metrics. \textbf{Bold} values indicate the higher score between the baseline and fine-tuned models for each metric, domain, and language pair. Arrows indicate significantly worse ($\downarrow$) and better ($\uparrow$) performance compared to the baseline according to each metric, domain, and language pair (p-value < 0.05).}
\label{tab:en_to_es_pt}
\end{table*}
\begin{table*}[ht]
\begin{minipage}[b]{0.5\textwidth}
\centering
\begin{tabular}{l|rrr}
\hline
\multirow{2}{*}{} & \multicolumn{3}{c}{en→it} \\
 & \multicolumn{1}{l}{BLEU} & \multicolumn{1}{l}{COMET} & \multicolumn{1}{l}{CometKiwi} \\ \hline
med\_base & 29.7 & 0.8481 & 0.8392 \\
sub2med & 28.6$\downarrow$ & 0.8444$\downarrow$ & 0.8351$\downarrow$ \\
ted2med & 29.4 & 0.8477 & 0.8416 \\
med\_ft & \textbf{30.6}$\uparrow$ & \textbf{0.8563}$\uparrow$ & \textbf{0.8456}$\uparrow$ \\ \hline
leg\_base & 36.9 & 0.8721 & 0.8567 \\
sub2leg & 35.1$\downarrow$ & 0.8645$\downarrow$ & 0.8492$\downarrow$ \\
ted2leg & 36.6 & 0.8717 & 0.8657$\uparrow$ \\
leg\_ft & \textbf{38.3}$\uparrow$ & \textbf{0.8827}$\uparrow$ & \textbf{0.8714}$\uparrow$ \\ \hline
it\_base & 29.2 & 0.7964 & 0.7753 \\
sub2it & 29.9 & 0.7962 & 0.7798 \\
ted2it & 30.8$\uparrow$ & 0.8023$\uparrow$ & 0.7909$\uparrow$ \\
it\_ft & \textbf{33.4}$\uparrow$ & \textbf{0.8151}$\uparrow$ & \textbf{0.7974}$\uparrow$ \\ \hline\hline
sub\_base & 17.8 & 0.7446 & 0.7783 \\
sub\_ft & \textbf{18.8}$\uparrow$ & \textbf{0.7532}$\uparrow$ & \textbf{0.7845}$\uparrow$ \\ \hline
ted\_base & 29.3 & 0.814 & 0.8141 \\
ted\_ft & \textbf{29.6} & \textbf{0.8199}$\uparrow$ & \textbf{0.8176}$\uparrow$ \\ \hline\hline
gen\_base & \textbf{31.5} & \textbf{0.8004} & \textbf{0.7837} \\
gen\_ft & 31.3 & 0.8002 & 0.7821 \\ \hline
\end{tabular}
\end{minipage}
\hfill
\begin{minipage}[b]{0.5\textwidth}
\centering
\begin{tabular}{l|rrr}
\hline
\multirow{2}{*}{} & \multicolumn{3}{c}{en→fr} \\
 & \multicolumn{1}{l}{BLEU} & \multicolumn{1}{l}{COMET} & \multicolumn{1}{l}{CometKiwi} \\ \hline
med\_base & 28.6 & 0.8179 & 0.8397 \\
sub2med & 27.9$\downarrow$ & 0.8133$\downarrow$ & 0.8346$\downarrow$ \\
ted2med & 28$\downarrow$ & 0.8155 & 0.8394 \\
med\_ft & \textbf{29.2}$\uparrow$ & \textbf{0.8293}$\uparrow$ & \textbf{0.8462}$\uparrow$ \\ \hline
leg\_base & 41.8 & 0.8533 & 0.8513 \\
sub2leg & 40.2$\downarrow$ & 0.8514 & 0.8483$\downarrow$ \\
ted2leg & 42 & 0.8589$\uparrow$ & 0.8597$\uparrow$ \\
leg\_ft & \textbf{43.7}$\uparrow$ & \textbf{0.8694}$\uparrow$ & \textbf{0.8668}$\uparrow$ \\ \hline
it\_base & 25.2 & 0.7386 & 0.7734 \\
sub2it & 24.7 & 0.7335$\downarrow$ & 0.7739 \\
ted2it & 26.5$\uparrow$ & 0.7451$\uparrow$ & 0.7871$\uparrow$ \\
it\_ft & \textbf{29.5}$\uparrow$ & \textbf{0.7619}$\uparrow$ & \textbf{0.7939}$\uparrow$ \\ \hline\hline
sub\_base & 16.1 & 0.6988 & 0.7759 \\
sub\_ft & \textbf{17.5}$\uparrow$ & \textbf{0.7073}$\uparrow$ & \textbf{0.7809}$\uparrow$ \\ \hline
ted\_base & 33.7 & 0.7926 & 0.81 \\
ted\_ft & \textbf{35}$\uparrow$ & \textbf{0.7955} & \textbf{0.8124} \\ \hline\hline
gen\_base & \textbf{26.1} & \textbf{0.7534} & 0.7829 \\
gen\_ft & 25.1$\downarrow$ & 0.753 & \textbf{0.7835} \\ \hline
\end{tabular}
\end{minipage}
\caption{Translation performance on en→it and en→fr language pairs for all models, evaluated using BLEU, COMET, and CometKiwi metrics. \textbf{Bold} values indicate the higher score between the baseline and fine-tuned models for each metric, domain, and language pair. Arrows indicate significantly worse ($\downarrow$) and better ($\uparrow$) performance compared to the baseline according to each metric, domain, and language pair (p-value < 0.05).}
\label{tab:en_to_it_fr}
\end{table*}
\begin{table*}[ht]
\begin{minipage}[b]{0.5\textwidth} \centering
\begin{tabular}{l|rrr}
\hline
\multirow{2}{*}{} & \multicolumn{3}{c}{en→cs} \\
 & \multicolumn{1}{l}{BLEU} & \multicolumn{1}{l}{COMET} & \multicolumn{1}{l}{CometKiwi} \\ \hline
med\_base & \textbf{22.6} & 0.8582 & 0.8313 \\
sub2med & 21.6$\downarrow$ & 0.8506$\downarrow$ & 0.8249$\downarrow$ \\
ted2med & 22.1 & 0.8595 & 0.8335 \\
med\_ft & 22.5 & \textbf{0.8659}$\uparrow$ & \textbf{0.8402}$\uparrow$ \\ \hline
leg\_base & 30.6 & 0.8905 & 0.8409 \\
sub2leg & 29.6$\downarrow$ & 0.8866$\downarrow$ & 0.8394 \\
ted2leg & 31.3$\uparrow$ & 0.8923 & 0.8582$\uparrow$ \\
leg\_ft & \textbf{33.4}$\uparrow$ & \textbf{0.9038}$\uparrow$ & \textbf{0.8614}$\uparrow$ \\ \hline
it\_base & 19.1 & 0.7872 & 0.7558 \\
sub2it & 18$\downarrow$ & 0.7829 & 0.7561 \\
ted2it & \textbf{19.7} & 0.7942$\uparrow$ & 0.7709$\uparrow$ \\
it\_ft & \textbf{19.7} & \textbf{0.8147}$\uparrow$ & \textbf{0.7855}$\uparrow$ \\ \hline\hline
sub\_base & 15 & 0.7594 & 0.7622 \\
sub\_ft & \textbf{15.3} & \textbf{0.766}$\uparrow$ & \textbf{0.7644} \\ \hline
ted\_base & 21.8 & 0.8082 & 0.79 \\ 
ted\_ft & \textbf{21.9} & \textbf{0.8194}$\uparrow$ & \textbf{0.7997}$\uparrow$ \\ \hline\hline
gen\_base & \textbf{25.1} & 0.7831 & 0.7471 \\
gen\_ft & 24.7 & \textbf{0.7874} & \textbf{0.7502} \\ \hline
\end{tabular}
\end{minipage}
\hfill
\begin{minipage}[b]{0.5\textwidth} \centering
\begin{tabular}{l|rrr}
\hline
\multirow{2}{*}{} & \multicolumn{3}{c}{en→pl} \\
 & \multicolumn{1}{l}{BLEU} & \multicolumn{1}{l}{COMET} & \multicolumn{1}{l}{CometKiwi} \\ \hline
med\_base & 18 & 0.8234 & 0.8063 \\
sub2med & 16.9$\downarrow$ & 0.815$\downarrow$ & 0.7987$\downarrow$ \\
ted2med & 17.4$\downarrow$ & 0.8217 & 0.807 \\
med\_ft & \textbf{18.1} & \textbf{0.8342}$\uparrow$ & \textbf{0.817}$\uparrow$ \\ \hline
leg\_base & 33.2 & 0.8796 & 0.8251 \\
sub2leg & 30.7$\downarrow$ & 0.8729$\downarrow$ & 0.8201$\downarrow$ \\
ted2leg & 32.8 & 0.8823 & 0.8358$\uparrow$ \\
leg\_ft & \textbf{34.2}$\uparrow$ & \textbf{0.8906}$\uparrow$ & \textbf{0.8427}$\uparrow$ \\ \hline
it\_base & 21.2 & 0.7818 & 0.7518 \\
sub2it & 20.3$\downarrow$ & 0.7793 & 0.7488 \\
ted2it & 21.7 & 0.7898$\uparrow$ & 0.7629$\uparrow$ \\
it\_ft & \textbf{22.5}$\uparrow$ & \textbf{0.8078}$\uparrow$ & \textbf{0.776}$\uparrow$ \\ \hline\hline
sub\_base & 14.6 & 0.7472 & 0.755 \\
sub\_ft & \textbf{15.1} & \textbf{0.7566}$\uparrow$ & \textbf{0.7576} \\ \hline
ted\_base & 16.2 & 0.7857 & 0.775 \\
ted\_ft & \textbf{16.6}$\uparrow$ & \textbf{0.7934}$\uparrow$ & \textbf{0.7819}$\uparrow$ \\ \hline\hline
gen\_base & \textbf{18.4} & \textbf{0.763} & 0.7445 \\
gen\_ft & 18.1 & 0.7604 & \textbf{0.7452} \\ \hline
\end{tabular}
\end{minipage}
\caption{Translation performance on en→cs and en→pl language pairs for all models, evaluated using BLEU, COMET, and CometKiwi metrics. \textbf{Bold} values indicate the higher score between the baseline and fine-tuned models for each metric, domain, and language pair. Arrows indicate significantly worse ($\downarrow$) and better ($\uparrow$) performance compared to the baseline according to each metric, domain, and language pair (p-value < 0.05).}
\label{tab:en_to_cs_pl}
\end{table*}
\begin{table*}[ht]
\centering
\begin{tabular}{l|rrr}
\hline
\multirow{2}{*}{} & \multicolumn{3}{c}{en→el} \\
 & \multicolumn{1}{l}{BLEU} & \multicolumn{1}{l}{COMET} & \multicolumn{1}{l}{CometKiwi} \\ \hline
med\_base & \textbf{29.1} & 0.8595 & 0.8152 \\
sub2med & 27.4$\downarrow$ & 0.8491$\downarrow$ & 0.8039$\downarrow$ \\
ted2med & 28.4$\downarrow$ & 0.8585 & 0.8152 \\
med\_ft & 28.8 & \textbf{0.8644}$\uparrow$ & \textbf{0.8214}$\uparrow$ \\ \hline
leg\_base & 40.7 & 0.8972 & 0.8384 \\
sub2leg & 38.4$\downarrow$ & 0.8919$\downarrow$ & 0.835 \\
ted2leg & 41.3$\uparrow$ & 0.899 & 0.8499$\uparrow$ \\
leg\_ft & \textbf{43.9}$\uparrow$ & \textbf{0.9059}$\uparrow$ & \textbf{0.8552}$\uparrow$ \\ \hline
it\_base & 27.1 & 0.7972 & 0.7563 \\
sub2it & 25.7$\downarrow$ & 0.7826$\downarrow$ & 0.7469$\downarrow$ \\
ted2it & 28.3$\uparrow$ & 0.8009 & 0.7662$\uparrow$ \\
it\_ft & \textbf{28.5}$\uparrow$ & \textbf{0.816}$\uparrow$ & \textbf{0.7754}$\uparrow$ \\ \hline\hline
sub\_base & \textbf{15.7} & \textbf{0.7768} & \textbf{0.7695} \\
sub\_ft & 15.6 & 0.7725 & 0.7651 \\ \hline
ted\_base & 29 & 0.8439 & 0.7934 \\
ted\_ft & \textbf{29.2} & \textbf{0.8475}$\uparrow$ & \textbf{0.7991}$\uparrow$ \\ \hline\hline
gen\_base & \textbf{26.2} & \textbf{0.8138} & \textbf{0.7754} \\
gen\_ft & 25.4$\downarrow$ & 0.8065$\downarrow$ & 0.7724 \\ \hline
\end{tabular}
\caption{Translation performance on en→el language pair for all models, evaluated using BLEU, COMET, and CometKiwi metrics. \textbf{Bold} values indicate the higher score between the baseline and fine-tuned models for each metric and domain. Arrows indicate significantly worse ($\downarrow$) and better ($\uparrow$) performance compared to the baseline according to each metric and domain (p-value < 0.05).}
\label{tab:en_to_el}
\end{table*}

\end{document}